\documentclass[11pt,a4paper]{article}
\usepackage[hyperref]{emnlp-ijcnlp-2019}
\usepackage{times}
\usepackage{latexsym}
\usepackage{color}
\usepackage{booktabs}
\usepackage{graphicx}
\usepackage{comment}
\usepackage{epsfig}
\usepackage[aboveskip=-2pt]{subcaption}

\usepackage{url}
\usepackage{linguex}

\aclfinalcopy 

\title{Quantity doesn't buy quality syntax with neural language models}

\author{Marten van Schijndel\\
  Cornell University \\
  {\tt mv443@cornell.edu} \\\And
  Aaron Mueller \\
  Johns Hopkins University\\
  {\tt amueller@jhu.edu} \\\And
  Tal Linzen \\
  Johns Hopkins University \\
  {\tt tal.linzen@jhu.edu}}

\date{}

\begin{document}
\maketitle
\begin{abstract}
Recurrent neural networks can learn to predict upcoming words remarkably well on average; in syntactically complex contexts, however, they often assign unexpectedly high probabilities to ungrammatical words. We investigate to what extent these shortcomings can be mitigated by increasing the size of the network and the corpus on which it is trained. We find that gains from increasing network size are minimal beyond a certain point. Likewise, expanding the training corpus yields diminishing returns; we estimate that the training corpus would need to be unrealistically large for the models to match human performance. A comparison to GPT and BERT, Transformer-based models trained on billions of words, reveals that these models perform even more poorly than our LSTMs in some constructions. Our results make the case for more data efficient architectures.
\end{abstract}

\section{Introduction}

\setlength{\Exlabelwidth}{0.25em}
\setlength{\SubExleftmargin}{1.35em}

Recurrent neural network language models (LMs) can learn to predict upcoming words with remarkably low perplexity \cite{mikolov2010recurrent,jozefowicz2016exploring,radfordetal19}. This overall success has motivated targeted paradigms that measure whether the LM's predictions reflect a correct analysis of sentence structure. One such evaluation strategy compares the probability assigned by the LM to a minimal pair of sentences differing only in grammaticality \cite{linzenetal16}. In the following example, the LM is expected to assign a higher probability to the sentence when the verb agrees in number with the subject \ref{ex:simple_agreement_grammatical} than when it does not \ref{ex:simple_agreement_ungrammatical}:

\ex.\a.The author \underline{laughs}.\label{ex:simple_agreement_grammatical}
    \b.*The author \underline{laugh}.\label{ex:simple_agreement_ungrammatical}

    RNN LMs have been shown to favor the grammatical variant in the vast majority of cases sampled at random from a corpus \cite{linzenetal16}, but their accuracy decreases in the presence of distracting nouns intervening between the head of the subject and the verb, especially when those nouns are in relative clauses \cite{marvinlinzen18}.
Can we hope to address these deficits by training larger and larger networks on larger and larger corpora, relying on the ``unreasonable effectiveness'' of massive datasets \cite{halevy2009unreasonable} and computational power?\footnote{\url{http://www.incompleteideas.net/IncIdeas/BitterLesson.html}} Or would architectural advances be necessary to improve our LMs' syntactic representations \cite{kuncoroetal18}?

This paper takes a first step towards addressing this question. We train 125 RNN LMs with long short-term memory \citep[LSTM,][]{hochreiterschmidhuber97} units, systematically varying the size of the training corpus and the dimensionality of the models' hidden layer, and track the relationship between these parameters and the performance of the models on agreement dependencies in a range of syntactic constructions \cite{marvinlinzen18}. We also compare our RNNs' accuracy to that of GPT \cite{radfordetal18} and BERT \cite{devlinetal19}, Transformer-based LMs trained on very large corpora.

We find that model capacity does not consistently improve performance beyond a minimum threshold. Increased corpus size likewise has a moderate and inconsistent effect on accuracy. We estimate that even if training data yielded consistent improvements, an unreasonable amount of data would be required to match human accuracy. We conclude that reliable and data-efficient learning of syntax is likely to require external supervision signals or a stronger inductive bias than that provided by RNNs and Transformers.

\section{Language models}
\paragraph{Architecture:} All of the models we trained consisted of two LSTM layers. We trained models with 100, 200, 400, 800 or 1600 units in each hidden layer. Input and output weights were tied during training \cite{presswolf17,inanetal17}; consequently, the input embedding had the same dimensionality as the hidden layers.

\paragraph{Training data:}
We trained networks of each size on 2M, 10M, 20M, 40M and 80M words (2M~=~2 million). 
We extracted five disjoint sections of the WikiText-103 corpus \cite{merityetal16} for each corpus size;\footnote{We made each of the 40M- and 80M-token training sets as disjoint as possible, but since Wikitext-103 only contains 103M tokens, it was not possible to make them wholly disjoint using Wikitext-103 as the mother corpus.} in total, we trained 125 models (5 layer sizes $\times$ 5 corpus sizes $\times$ 5 corpus subsets).\footnote{Each model was initialized randomly and was trained to convergence with a dropout of 0.2 using a batch size of 20, backpropagating error for 35 observations. An initial learning rate of 20 was gradually annealed.} 
We used the WikiText-103 validation set for validation.

\paragraph{Vocabulary:} To ensure comparability across different models trained on different data, we used the same vocabulary for all the models we trained.
The vocabulary consisted of an intersection of the 400k word GloVe vocabulary \cite{penningtonetal14} with the 50k words used by GRNN (see below); the resulting vocabulary had 28,438 words.

\paragraph{GRNN:} We also report the syntactic performance of a publicly available LSTM LM \cite[henceforth GRNN]{gulordavaetal18}. This trained model has been the focus of a considerable amount of analysis work in the past year. The model has two layers of 650 units each, and was trained on 80M words.

\paragraph{Comparison with Transformers:}

Finally, we report results from two publicly available LMs based on non-recurrent self-attention \cite[Transformers;][]{vaswanietal17}: GPT \cite{radfordetal18} and BERT \cite{devlinetal19}. Both of these models have been argued to learn powerful syntactic representations \cite{goldberg19,wolfe19}. 
We compare our results to those reported by \citet{wolfe19} on a similar challenge set for these two Transformer models.%
\footnote{The comparison is not exact because the Transformers were evaluated based on the rank of the two target verbs given the prefix, and the LSTMs based on the total log-probability of the sentence (including the final period); in addition, \protect\citet{wolfe19} did not modify the dataset to use only high frequency verbs, as we describe in Section~\ref{sec:evaluation}.}

GPT is a 12-layer Transformer with 110~million parameters (compared to GRNN's 39~million parameters); it was trained on 1~billion words.
BERT has a similar architecture to GPT,%
\footnote{BERT Base showed more accurate syntactic predictions than BERT Large \protect\cite{goldberg19}, which has more parameters, so we only consider BERT Base.} with three differences: it is bidirectional, it was trained on 3.3 billion words, and it has a different training objective than the typical LM: it attempts to predict a single masked word in a sentence given the words both before and after the target word.
For comparability to the LSTMs and GPT, we examine the agreement performance of BERT when only the words before the target are given (in contrast to the bidirectional tests reported by \citealt{goldberg19}).

\section{Evaluation}
\label{sec:evaluation}

We tested each trained model on the constructions from the \newcite{marvinlinzen18} challenge set, which is based on the agreement paradigm described in the introduction.%
\footnote{\url{https://github.com/BeckyMarvin/LM_syneval}}
We replaced the verbs used by \citeauthor{marvinlinzen18} with the high-frequency verbs \emph{is/are}, \emph{was/were} and \emph{has/have}. This was done to ensure that even the models trained on smaller corpora will have had exposure to both forms of the verb in question.

We performed statistical tests of our hypotheses using Bayes factors, which quantify the support the data provide for a more complex model compared to a simpler one \cite{rouder2009bayesian}. We computed two-sample Bayes factors using {\tt ttestBF} from the {\tt BayesFactor} R package \cite{bayesfactor} using default settings. Our null hypothesis was that there is no difference in accuracy between the two sets of models in question (e.g., all models with 400 units per layer compared to all models with 800 units per layer). The magnitude of the resulting Bayes factor $K$ can be interpreted as follows \cite{jeffreys61}: $K < 1$ indicates that there is no difference in accuracy between the two model groups, and $K > 10$ provides strong evidence that the model groups obtain different accuracies. 

\section{Results}

\begin{table}
    \resizebox{\linewidth}{!}{
  \begin{tabular}{lrlr}
      \toprule
      \multicolumn{2}{c}{Corpus size} & \multicolumn{2}{c}{Layer size}\\
      \cmidrule(lr){1-2} \cmidrule(lr){3-4}
      2M $\rightarrow$ 10M & $\mathbf{5508.8}$ & 100 $\rightarrow$ 200 & $\mathbf{768.5}$\\
      10M $\rightarrow$ 20M & $0.1$ & 200 $\rightarrow$ 400 & $\mathbf{63.5}$\\
      20M $\rightarrow$ 40M & $\mathbf{12.9}$ & 400 $\rightarrow$ 800 & $0.2$\\
      40M $\rightarrow$ 80M & $0.2$ & 800 $\rightarrow$ 1600 & $0.1$\\
      \bottomrule
  \end{tabular}
  }
  \caption{Strength of evidence for improvements in agreement prediction accuracy as a result of increasing corpus size averaging across layer size (left) or layer size averaging across corpus size (right), as quantified by Bayes factors. Boldfaced Bayes factors indicate strong evidence of improvement.}\label{tab:genresults}
\end{table}

Increasing model size improved syntactic prediction accuracy up to 400 units per layer; further increases in model size had no effect (see Table~\ref{tab:genresults} for the statistical tests). Increasing the amount of training data impacted accuracy in an inconsistent way. Training on 10M tokens produced general improvements across all constructions compared to 2M tokens. Doubling the corpus to 20M did not affect accuracy, but doubling it again to 40M did. There was no evidence of further improvement between 40M and 80M words.  

\begin{figure*}[h!] 
  \begin{subfigure}[b]{0.485\textwidth}
  \includegraphics[width=\textwidth]{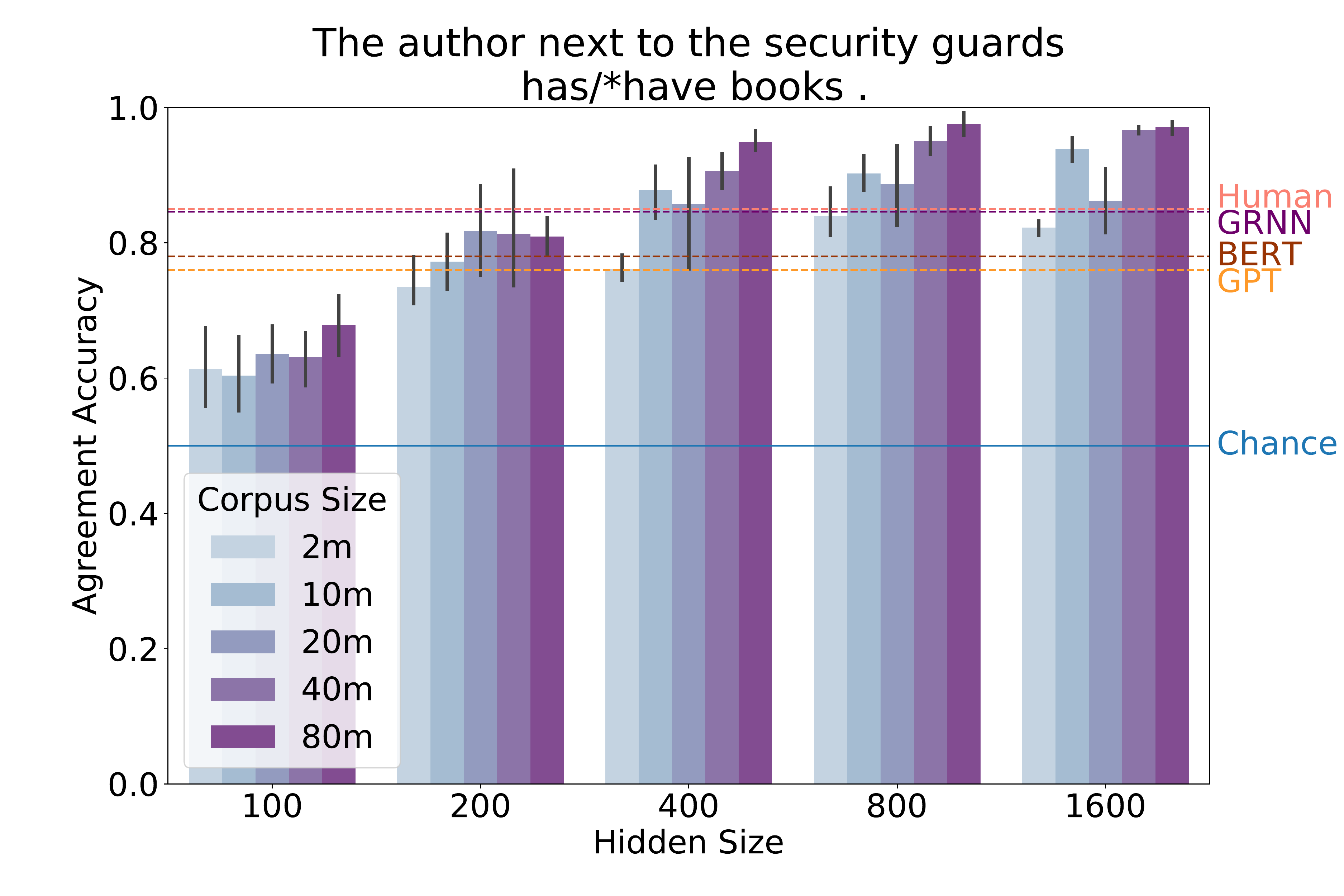}
  \caption{\label{fig:prep} Prepositional Phrase}
  \end{subfigure}\hfill
  \begin{subfigure}[b]{0.485\textwidth}
  \includegraphics[width=\textwidth]{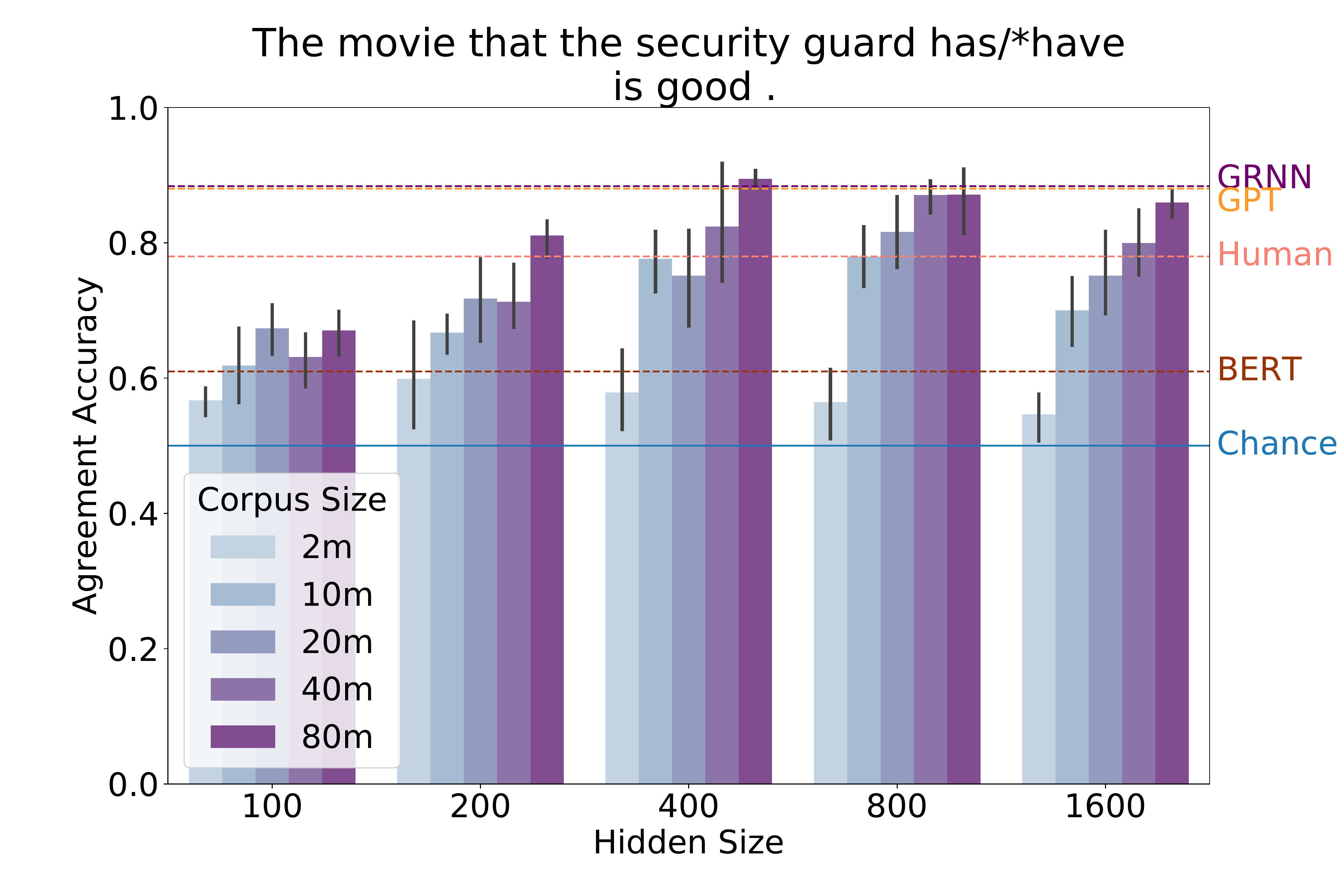}
  \caption{\label{fig:objrelwithin} Object Relative: Within}
  \end{subfigure}\\
  \begin{subfigure}[b]{0.485\textwidth}
  \includegraphics[width=\textwidth]{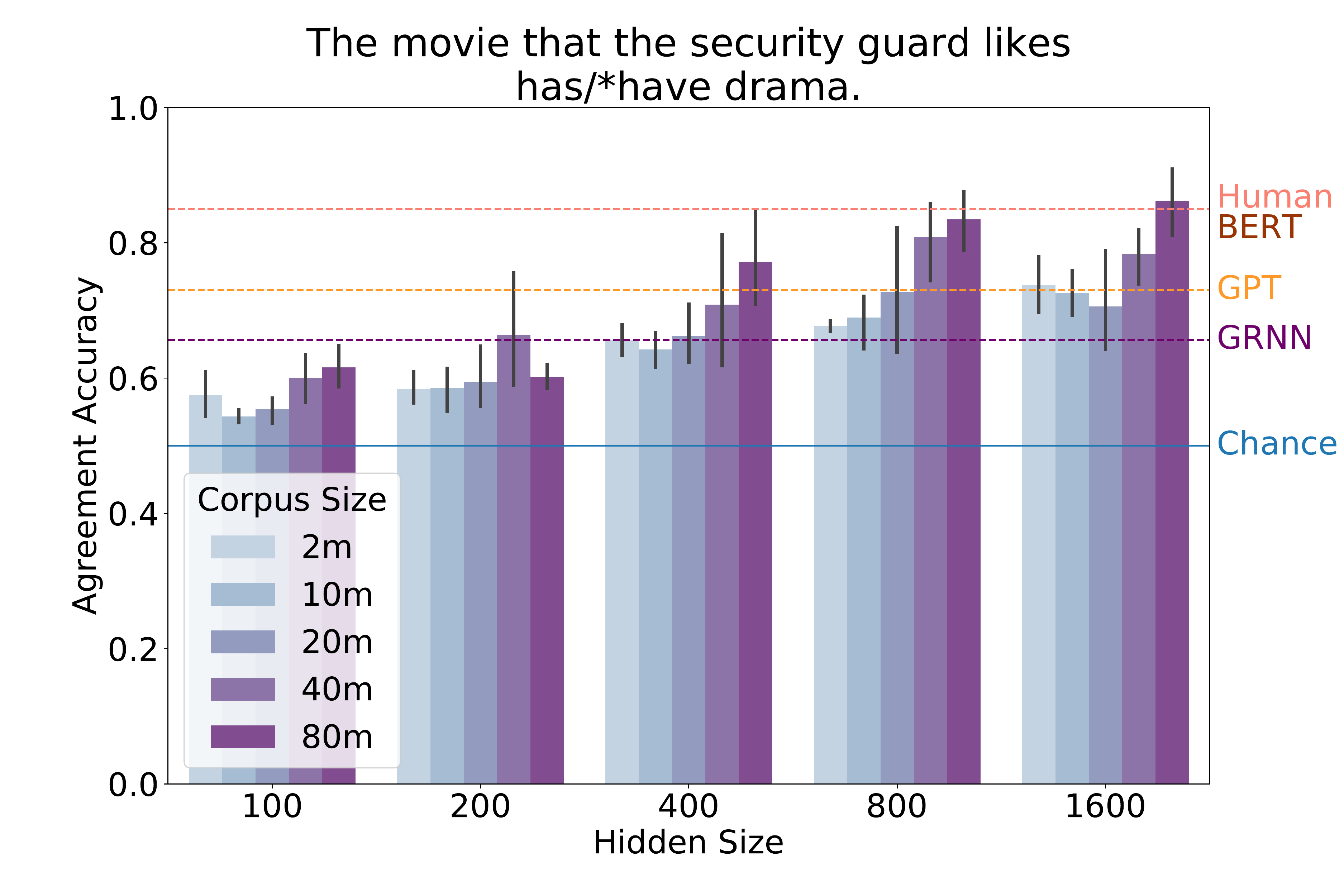}
  \caption{\label{fig:objrelacross} Object Relative: Across}
  \end{subfigure}\hfill
  \begin{subfigure}[b]{0.485\textwidth}
  \includegraphics[width=\textwidth]{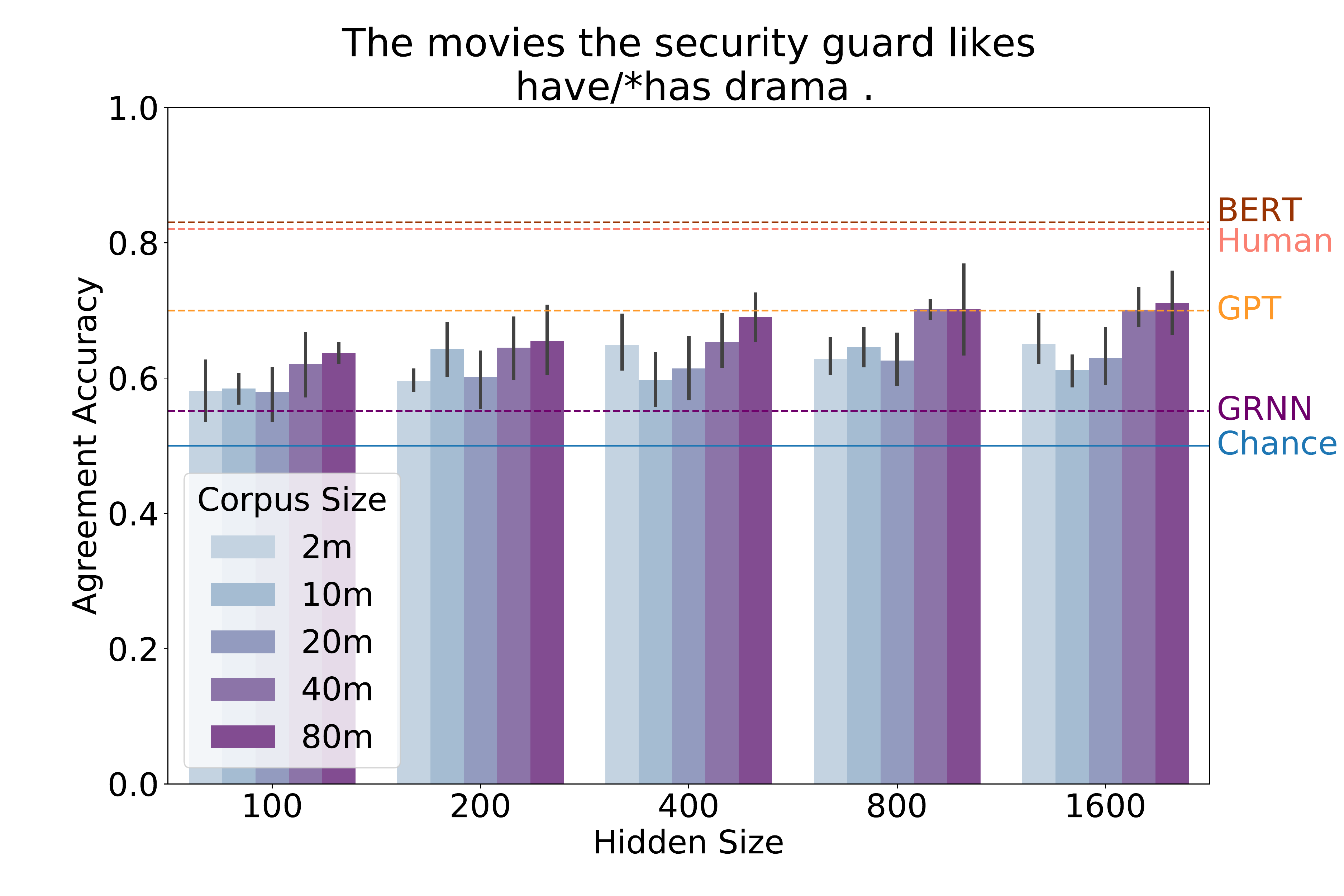}
  \caption{\label{fig:objrelacrossnocomp} Object Relative: Across (no that)}
  \end{subfigure}\\
  \begin{subfigure}[b]{0.485\textwidth}
  \includegraphics[width=\textwidth]{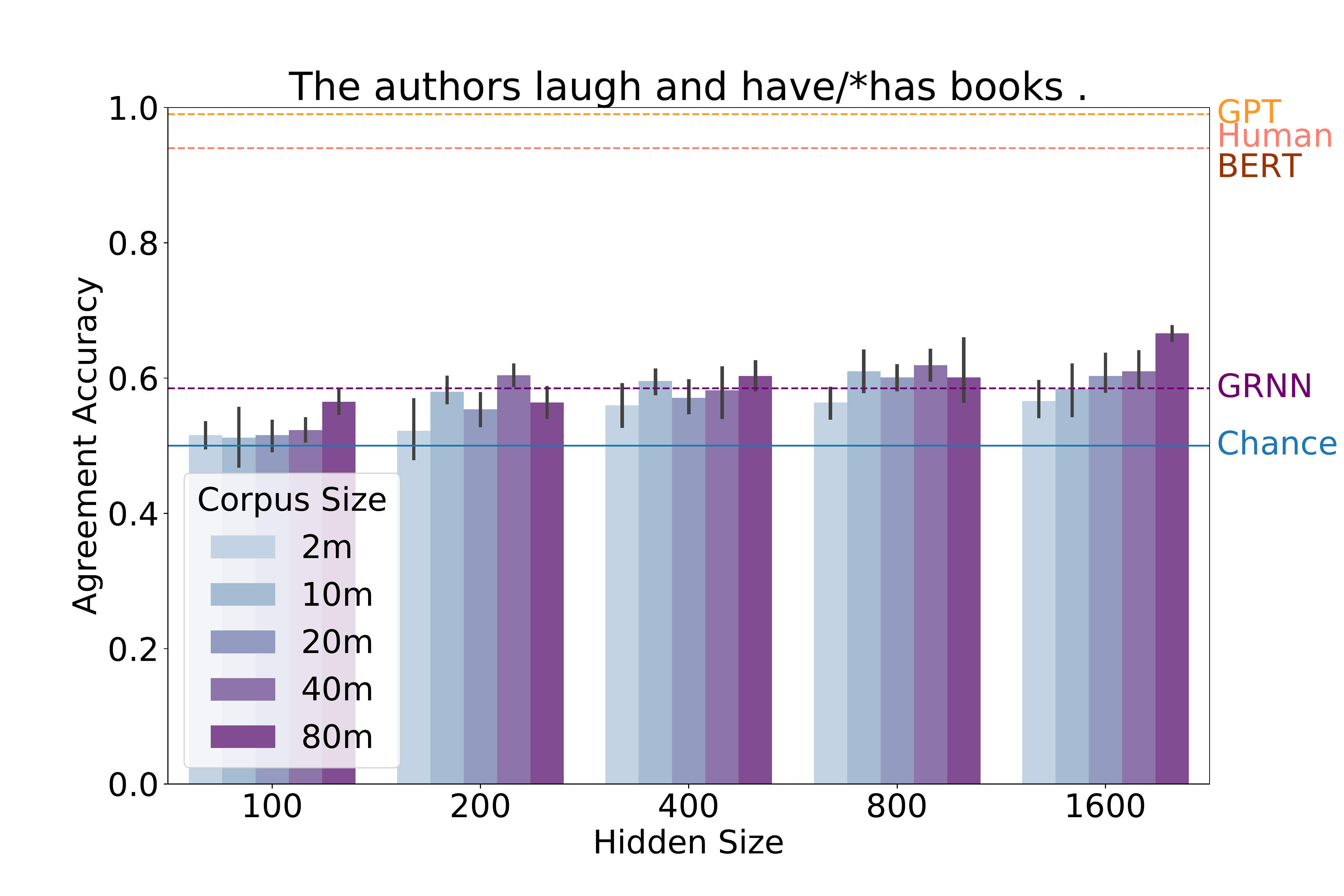}
  \caption{\label{fig:vpcoord} VP Coordination (Short)}
  \end{subfigure}\hfill
  \begin{subfigure}[b]{0.485\textwidth}
  \includegraphics[width=\textwidth]{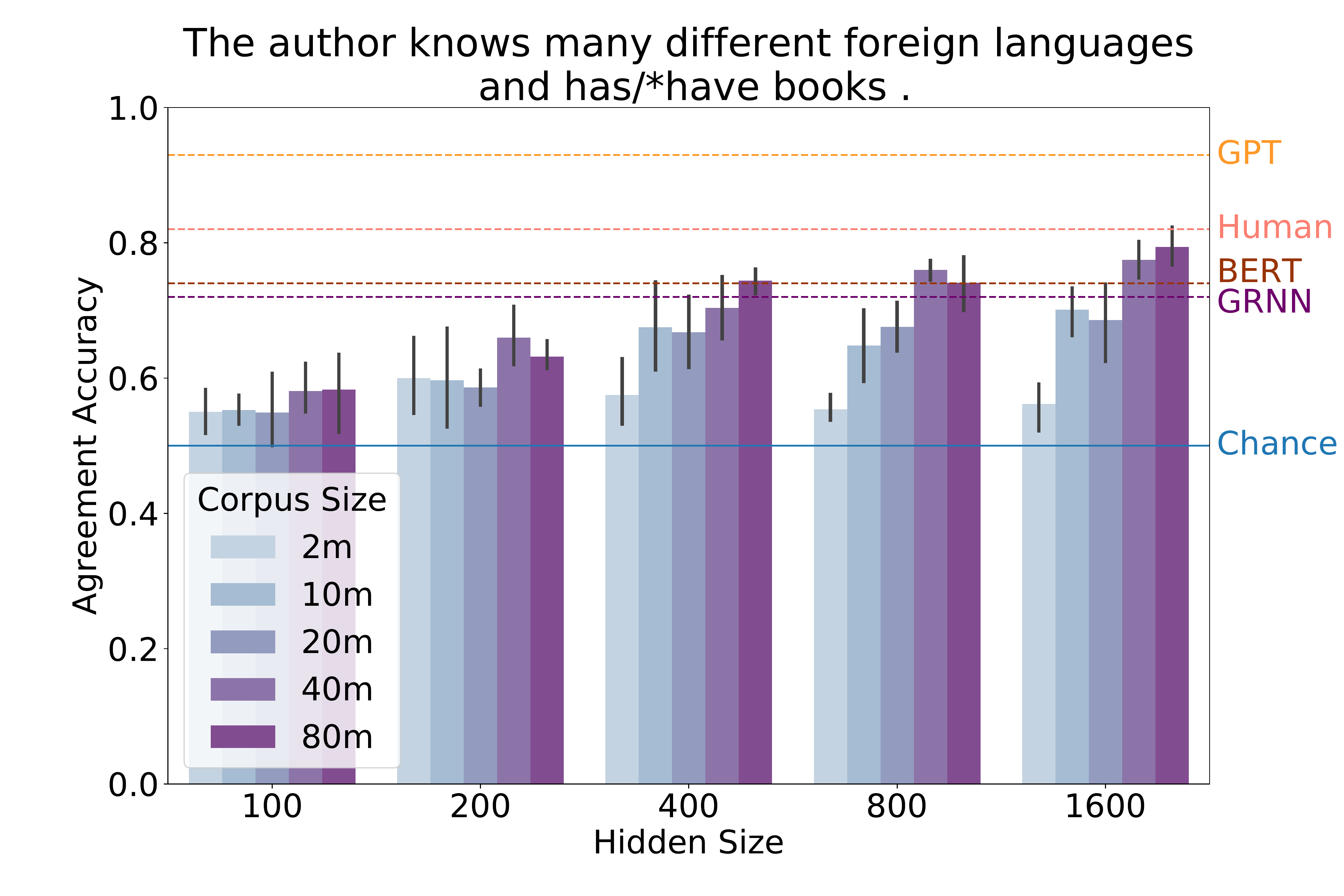}
  \caption{\label{fig:vpcoordlong} VP Coordination (Long)}
  \end{subfigure}\\
  \caption{LSTM agreement performance in several syntactic constructions. The solid horizontal line indicates chance performance. The dashed lines show the performance of GPT and BERT as reported by \protect\citet{wolfe19}, the performance of humans as reported by \protect\citet{marvinlinzen18}, and the performance of GRNN. Error bars reflect standard deviation across the five models in each category.}\label{fig:agreement}
\end{figure*}

In the remainder of this section we analyze the effect of increasing model size and training corpus size on the models' predictions for each construction in the data set.\footnote{See Table~\ref{tab:incremental} in the Appendix for a Bayes factor analysis of the improvement in each construction for each amount of training data.} A subset of the results is shown in Fig.~\ref{fig:agreement}; for the full results, see Figs.~\ref{fig:app1}, \ref{fig:app2} and \ref{fig:app3} in the Appendix.

\paragraph{Local number agreement:}
All models trained on 10M words or more obtained perfect or near-perfect accuracy (mean $>99$\%) in the cases where the verb was adjacent to its subject: simple agreement (\emph{the author has/*have books}) and agreement within a sentential complement (\emph{the mechanics said the author has/*have books}).
When trained on 2M words, the models performed slightly worse, but their accuracy was still very high (mean 95.6\%), regardless of model size. Overall, we conclude that the plurality of specific nouns and the generalization that a verb has to agree with a noun can be learned very quickly.

\paragraph{Attractors:} Agreement across subject relative clauses (\emph{the author that likes the guards has/*have books}) and across prepositional phrases (\emph{the author next to the guards has/*have books}; Fig.~\ref{fig:prep}) benefited from increasing the hidden layer size to 400, but showed little improvement when hidden layer size was increased further. Accuracy in these constructions consistently improved as the amount of training data increased. 

\paragraph{Object relative clauses:}

Expanding the training corpus improved local agreement \textbf{within} object relative clauses (\emph{the movies that the guard has/*have are good}; Fig.~\ref{fig:objrelwithin}) for all model sizes, but only improved agreement \textbf{across} those clauses (\emph{the movie that the guards like has/*have drama}; Fig.~\ref{fig:objrelacross}) in models with larger hidden layers. Larger hidden layers improved accuracy in object relatives only when a relativizer was present, and only up to about 80\% accuracy. When a sentence lacked an overt relativizer (\emph{the movies the security guard has/*have are good}; Fig.~\ref{fig:objrelacrossnocomp}), all models performed poorly, with accuracy levelling off around 70\%. 

\paragraph{Coordination:} Perhaps surprisingly, all LSTM LMs struggled with agreement in a coordinated verb phrase (\emph{the authors laugh and have/*has books}; Fig.~\ref{fig:vpcoord}), even though this construction does not include distracting nouns between the subject and the second verb. In larger models trained on more data, accuracy was higher when the second verb was \textbf{further} from the subject (\emph{the authors know many different languages and have/*has books}; Fig.~\ref{fig:vpcoordlong}). 

Training on more than 10M tokens did not improve accuracy in short VP coordination, even when the amount of data was multiplied by eight (10M $\rightarrow$ 80M: $K < 1$), unlike coordination across long VPs, which benefited from additional data (10M $\rightarrow$ 80M: $K > 90$). These results further challenge the assumption that increased amounts of training data will lead to adequately abstract syntactic representations: RNNs show a limited ability to generalize from instances of a construction that have longer constituents to instances with shorter constituents.

\paragraph{Reflexive anaphora:}

A reflexive pronoun such as \textit{themselves} must have an antecedent with the appropriate plurality in the same clause as the pronoun (\textit{The manager that the architects like doubted himself/*themselves}). Accuracy was not strongly affected by the parameters we varied: reflexive agreement accuracy across a relative clause was consistently mediocre (61\%--76\%) regardless of model size or the amount of training data.

\paragraph{Transformers:}
Despite having more parameters and having been trained on significantly larger corpora, the two Transformer models performed either as well as or more poorly than our LSTMs in seven of the ten subject-verb agreement conditions.
BERT underperformed GPT in several conditions despite having been trained on three times as many tokens as GPT.%
\footnote{\citet{goldberg19} reports much better results using a setup in which BERT has access to both left and right context. We hypothesize that the task is made significantly simpler when the model knows where the target word is relative to the end of the sentence. For example, if the point of prediction is at the last word of the sentence, it is also the last point at which the verb agreeing with the main clause subject could possibly occur; the model does not need to detect the end of the relative clause to perform the task in this case.}

    \begin{figure}
      \centering
      \begin{subfigure}[b]{0.24\textwidth}
        \includegraphics[width=\textwidth]{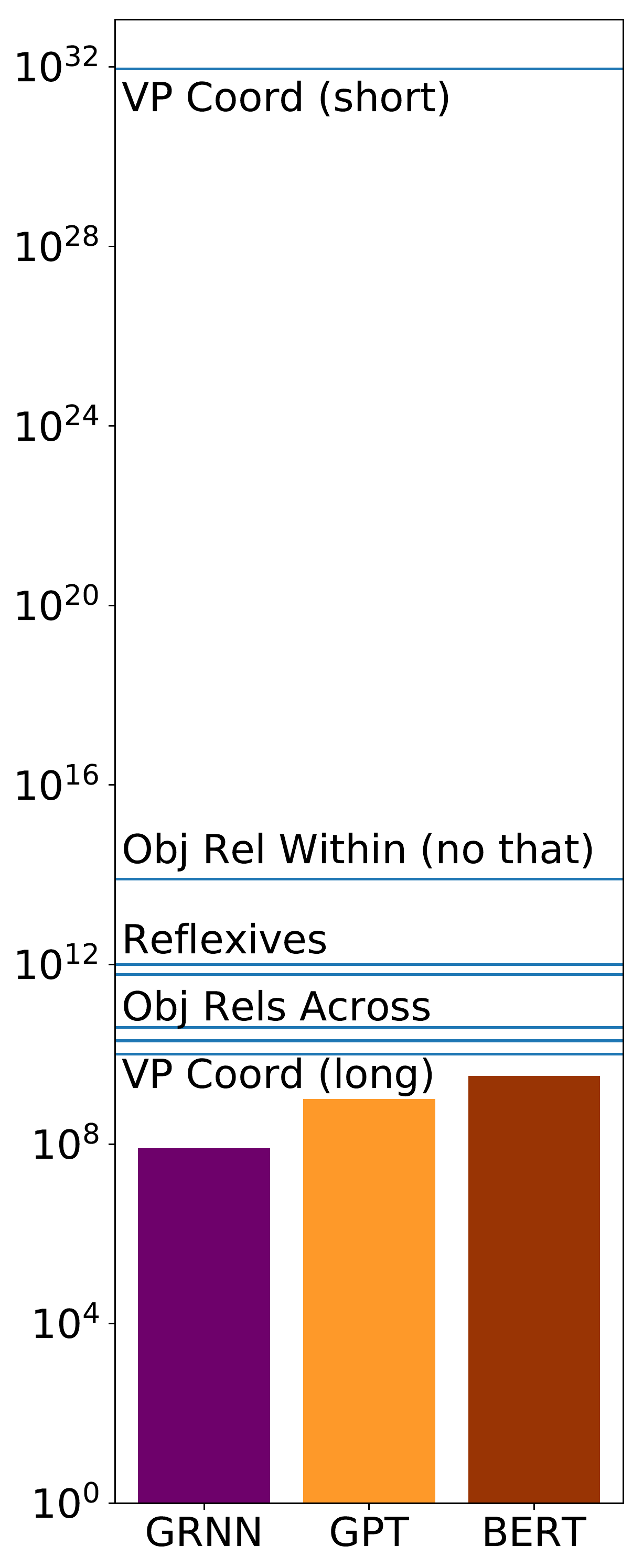}
        \caption{\label{fig:human} Human-like}
      \end{subfigure}\hfill
      \begin{subfigure}[b]{0.24\textwidth}
        \includegraphics[width=\textwidth]{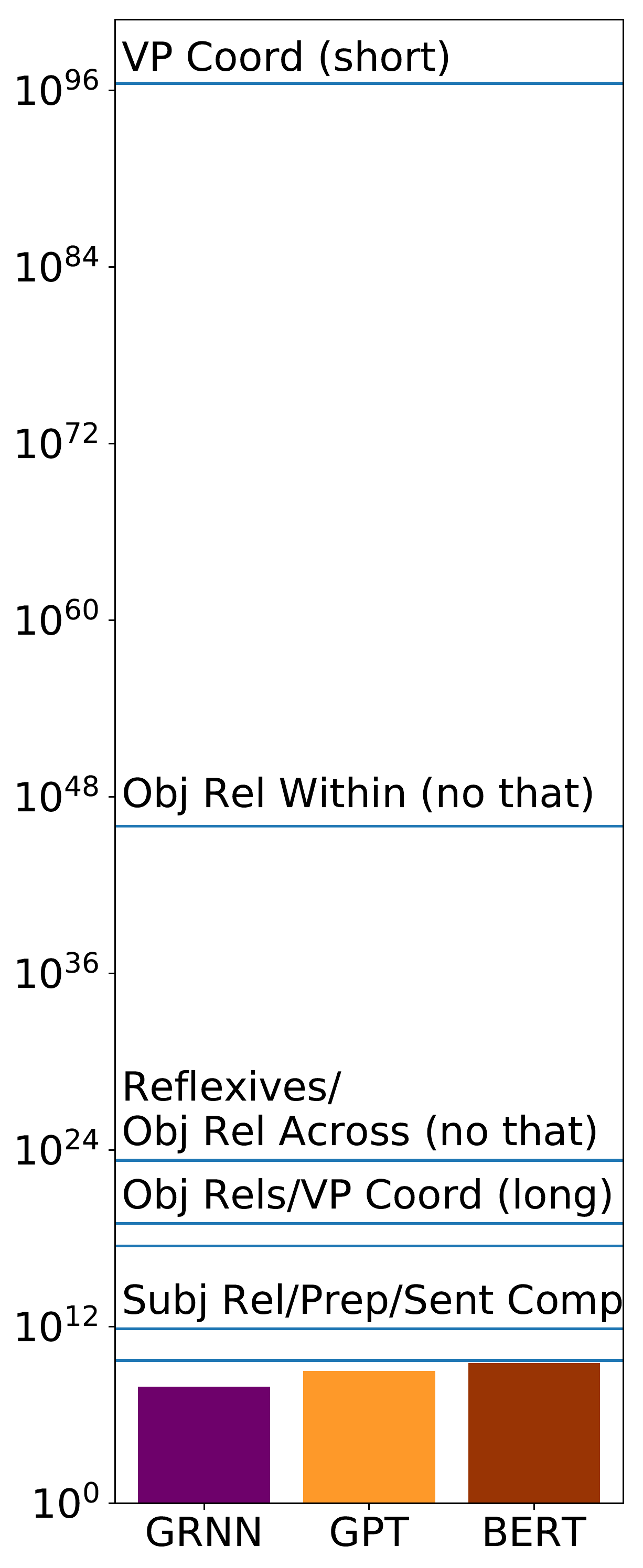}
        \caption{\label{fig:perfect} 99.99\%}
      \end{subfigure}
      \caption{Lines depict number of training tokens needed for LSTMs to achieve human-like (left) or 99.99\% accuracy (right) in each syntactic agreement condition, according to our estimates. Bars depict the amount of data on which each model was trained.}\label{fig:enough}
    \end{figure}

\section{How much data would be enough?}

How much training data would be required for an LSTM LM to perform at a human level (as reported by \citealt{marvinlinzen18}) in the conditions in which our models do not already perform at a human level? As a conservative estimate, we measured the error reduction achieved by doubling the data from 20M to 40M tokens (the largest error reduction we observed beyond 2M $\rightarrow$ 10M).%
\footnote{See Tables~\ref{tab:appenough} and \ref{tab:appenoughperfect} in the Appendix for data requirements estimated from other error reduction rates.} Under the assumption that each subsequent doubling of training data would produce the same percent error reduction, we predicted the amount of data required to obtain human-like and 99.99\% accuracy (see Fig.~\ref{fig:enough}).\footnote{Human performance on this task is well known to be far from perfect, with error rates approaching 25\% in some contexts \cite{bock1991broken}. While modeling human errors is of considerable interest to cognitive scientists \cite{linzenleonard18}, we believe that in most applied contexts it is desirable for the model to make no errors at all.} We found that every remaining construction would require over 10 billion tokens to achieve human-like performance, and most would require trillions of tokens to achieve perfect accuracy -- an impractically large amount of training data, especially for these relatively simple syntactic phenomena. 

\section{Discussion}

We have investigated the effect of network size and training corpus size on the quality of the syntactic representations of LSTM LMs, as measured by agreement prediction accuracy. 
Increased model size had limited benefits; models with 400 hidden units performed significantly better than smaller models, but further increases in network size had no effect. The limited effect of network size is consistent with previous findings on sequence labeling tasks \cite{reimersgurevych17,greffetal17}. We have also shown that increasing the amount of training data is unlikely to result in human-like accuracy in all cases. 

We found a striking difference in agreement accuracy between short and long coordinated verb phrases: performance on short phrases was poorer. While RNNs are known to struggle with generalizing short patterns to longer sequences, this pattern constitutes a failure to generalize to \emph{shorter} sequences \citep[cf.][]{trasketal18}; techniques for improving longer distance dependency learning in LMs \citep[e.g.,][]{trinhetal18,daietal19} are unlikely to mitigate this deficit. This suggests that challenge sets should include materials that can be used to ascertain whether the model's syntactic representations are robust to syntactically irrelevant factors such as constituent length.

GPT and BERT, Transformer models trained on very large corpora, did not consistently outperform the LSTMs trained on several orders of magnitude less data. Other studies suggest that Transformer models suffer from similar problems as the LSTMs we have analyzed. BERT's agreement accuracy decreases as the subject becomes more distant from its verb \cite{baconregier19}. Dramatically increasing the pre-training corpus for a BERT-like model from 562M words to 18G words only leads to a modest improvement in its natural language inference accuracy, from 81.7\% to 82.3\% \cite{baevskietal19}. Overall, this body of results points to the limited data efficiency of standard RNNs and Transformers, and indicates that learning syntax from realistic amounts of data---in particular the amount of data available to humans when they learn language---may require syntactically structured architectures or explicit syntactic supervision \cite{Enguehard17,kuncoroetal18,kuncoro2019scalable,wilcox2019structural}.

\bibliography{syntax_impacts}
\bibliographystyle{acl_natbib}

\clearpage
\appendix

\begin{figure*}[h!]
  \begin{subfigure}[b]{0.5\textwidth}
  \includegraphics[width=\textwidth]{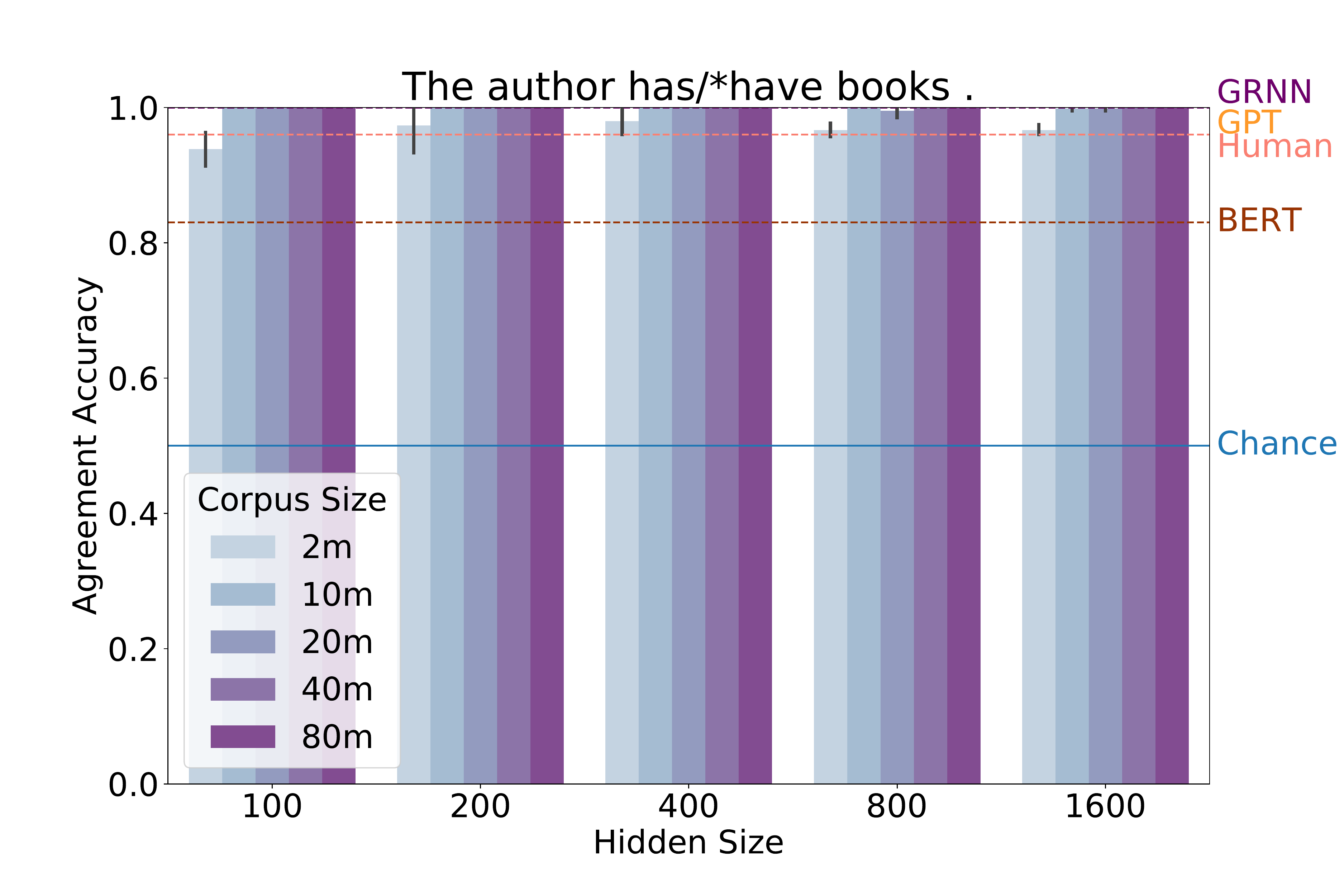}
  \caption{\label{fig:1} Simple Agreement}
  \end{subfigure}\hfill
  \begin{subfigure}[b]{0.5\textwidth}
  \includegraphics[width=\textwidth]{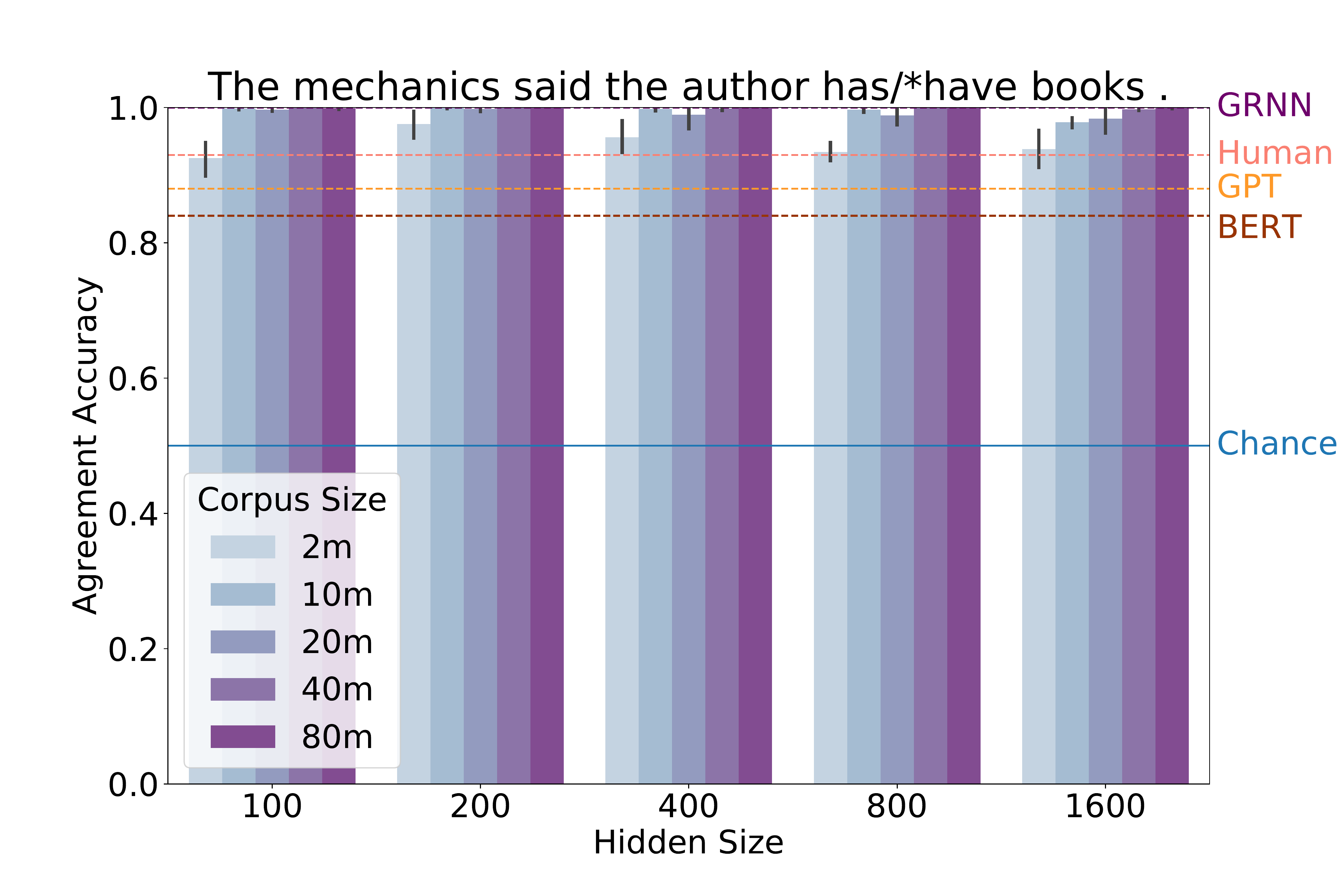}
  \caption{\label{fig:2} Sentential Complement}
  \end{subfigure}\\
  \begin{subfigure}[b]{0.5\textwidth}
  \includegraphics[width=\textwidth]{sizecmp_vp_coord.pdf}
  \caption{\label{fig:3} VP Coordination (Short)}
  \end{subfigure}\hfill
  \begin{subfigure}[b]{0.5\textwidth}
  \includegraphics[width=\textwidth]{sizecmp_long_vp_coord.pdf}
  \caption{\label{fig:4} VP Coordination (Long)}
  \end{subfigure}\\
  \begin{subfigure}[b]{0.5\textwidth}
  \includegraphics[width=\textwidth]{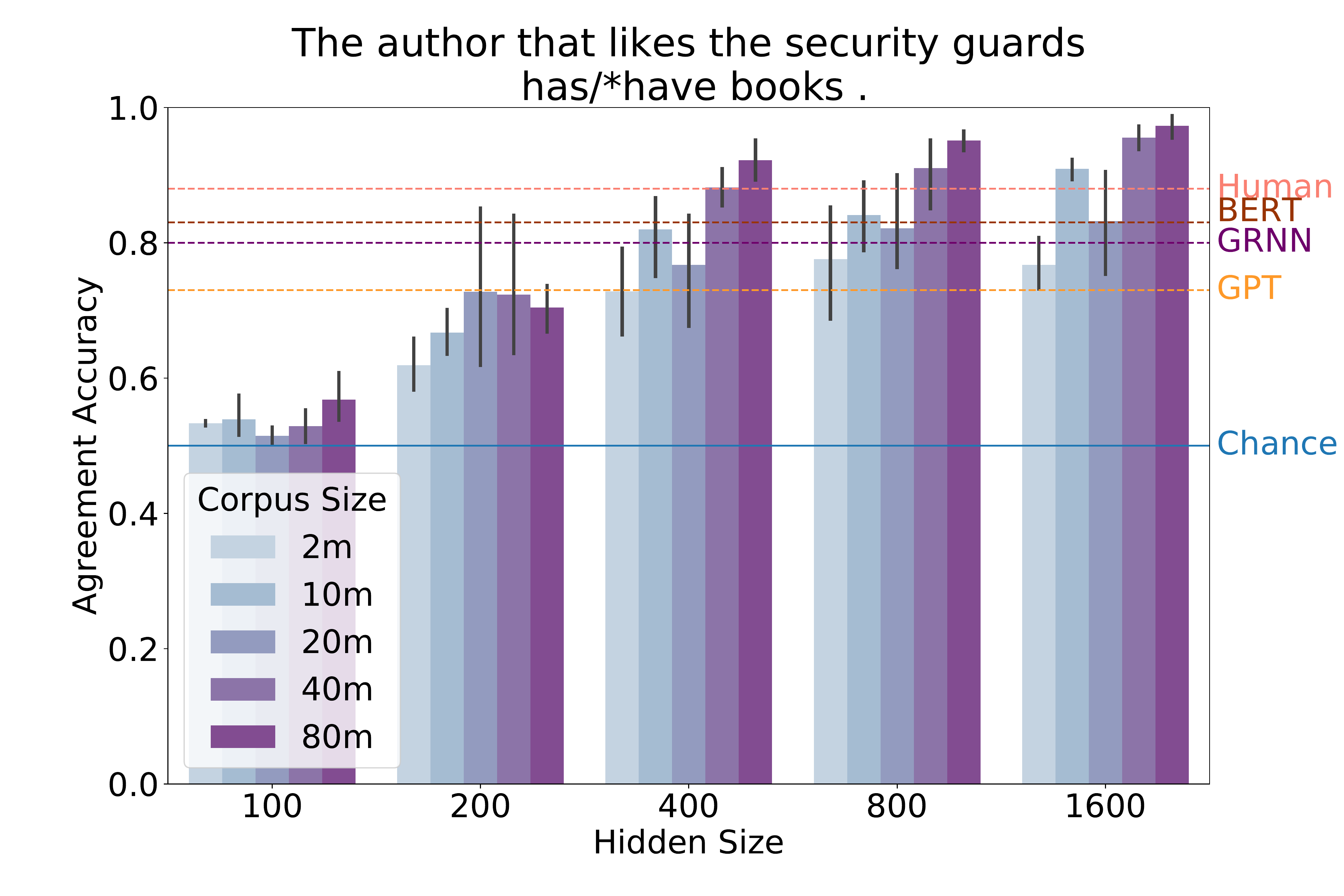}
  \caption{\label{fig:5} Subject Relative}
  \end{subfigure}\hfill
  \begin{subfigure}[b]{0.5\textwidth}
  \includegraphics[width=\textwidth]{sizecmp_prep.pdf}
  \caption{\label{fig:6} Prepositional Phrase}
  \end{subfigure}
\caption{Language model agreement performance when the verb is adjacent to its subject (\ref{fig:1},\ref{fig:2}), when the verb is coordinated with another verb (\ref{fig:3},\ref{fig:4}), when the verb and subject have an intervening relative clause (\ref{fig:5}), and when the verb and subject have an intervening prepositional phrase (\ref{fig:6}). The dashed horizontal lines show agreement performance of commonly-used large-scale models. Error bars reflect standard deviation across the five models in each category. GPT and BERT results are from those reported by \protect\citet{wolfe19}. Human results are those reported by \protect\citet{marvinlinzen18}.}\label{fig:app1}
\end{figure*}

\begin{figure*}
  \begin{subfigure}[b]{0.5\textwidth}
  \includegraphics[width=\textwidth]{sizecmp_obj_rel_within.pdf}
  \caption{\label{fig:7} Object Relative: Within}
  \end{subfigure}\hfill
  \begin{subfigure}[b]{0.5\textwidth}
  \includegraphics[width=\textwidth]{sizecmp_obj_rel_across.pdf}
  \caption{\label{fig:8} Object Relative: Across}
  \end{subfigure}\\
  \begin{subfigure}[b]{0.5\textwidth}
  \includegraphics[width=\textwidth]{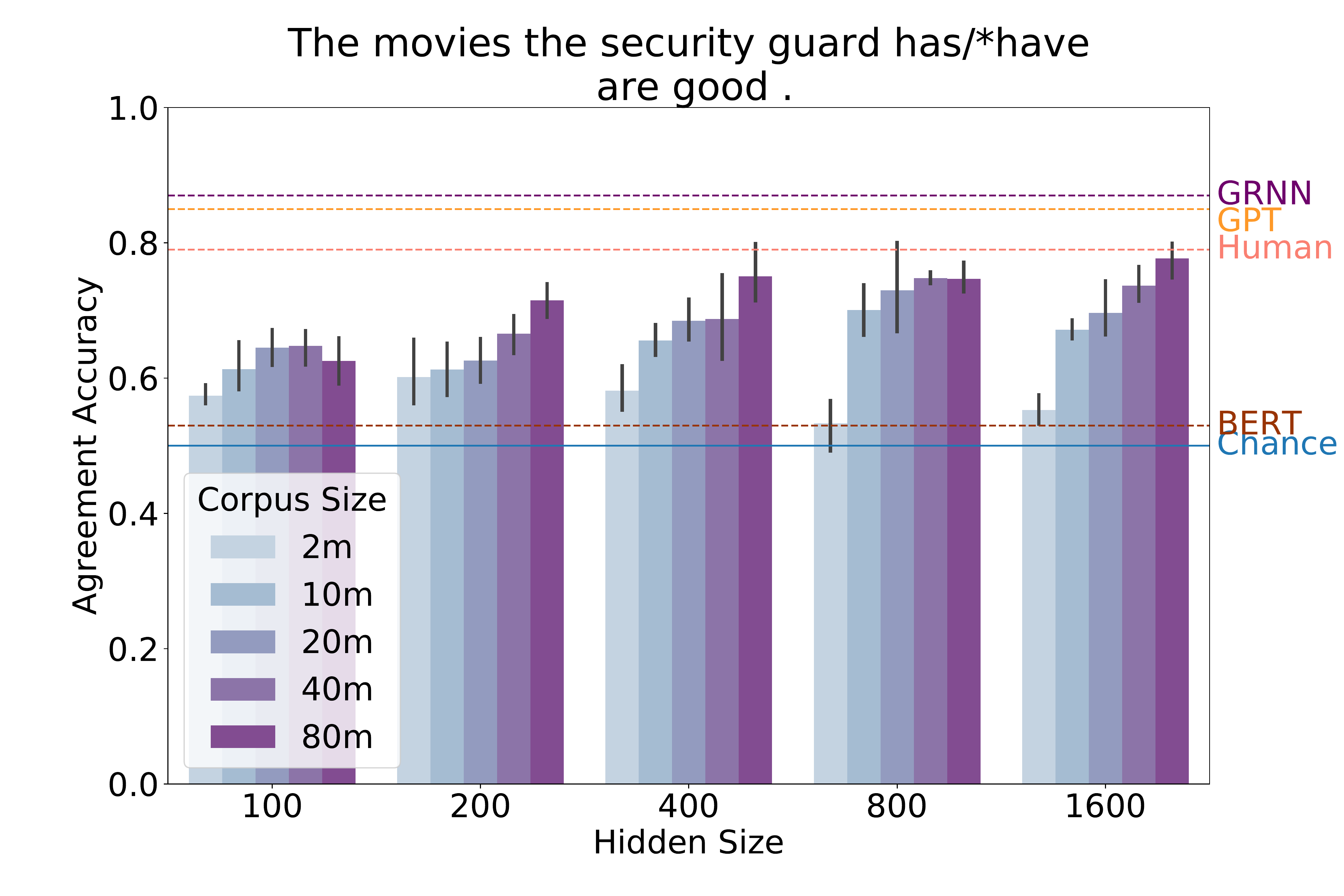}
  \caption{\label{fig:9} Object Relative: Within (no that)}
  \end{subfigure}\hfill
  \begin{subfigure}[b]{0.5\textwidth}
  \includegraphics[width=\textwidth]{sizecmp_obj_rel_no_comp_across.pdf}
  \caption{\label{fig:10} Object Relative: Across (no that)}
  \end{subfigure}
  \caption{Language model agreement performance when the target verb is \textbf{within} an object-modifying relative clause (Left) and when an object-modifying relative clause intervenes between the target verb and its subject (Right). These results distinguish between when the relative clause has an overt relativizer (Top) and when it lacks an overt relativizer (Bottom). The dashed horizontal lines show agreement performance of commonly-used large-scale models. Error bars reflect standard deviation across the five models in each category. GPT and BERT results are from those reported by \protect\citet{wolfe19}. Human results are those reported by \protect\citet{marvinlinzen18}.}\label{fig:app2}
\end{figure*}

\begin{figure*}
  \centering
  \begin{subfigure}[b]{0.5\textwidth}
  \includegraphics[width=\textwidth]{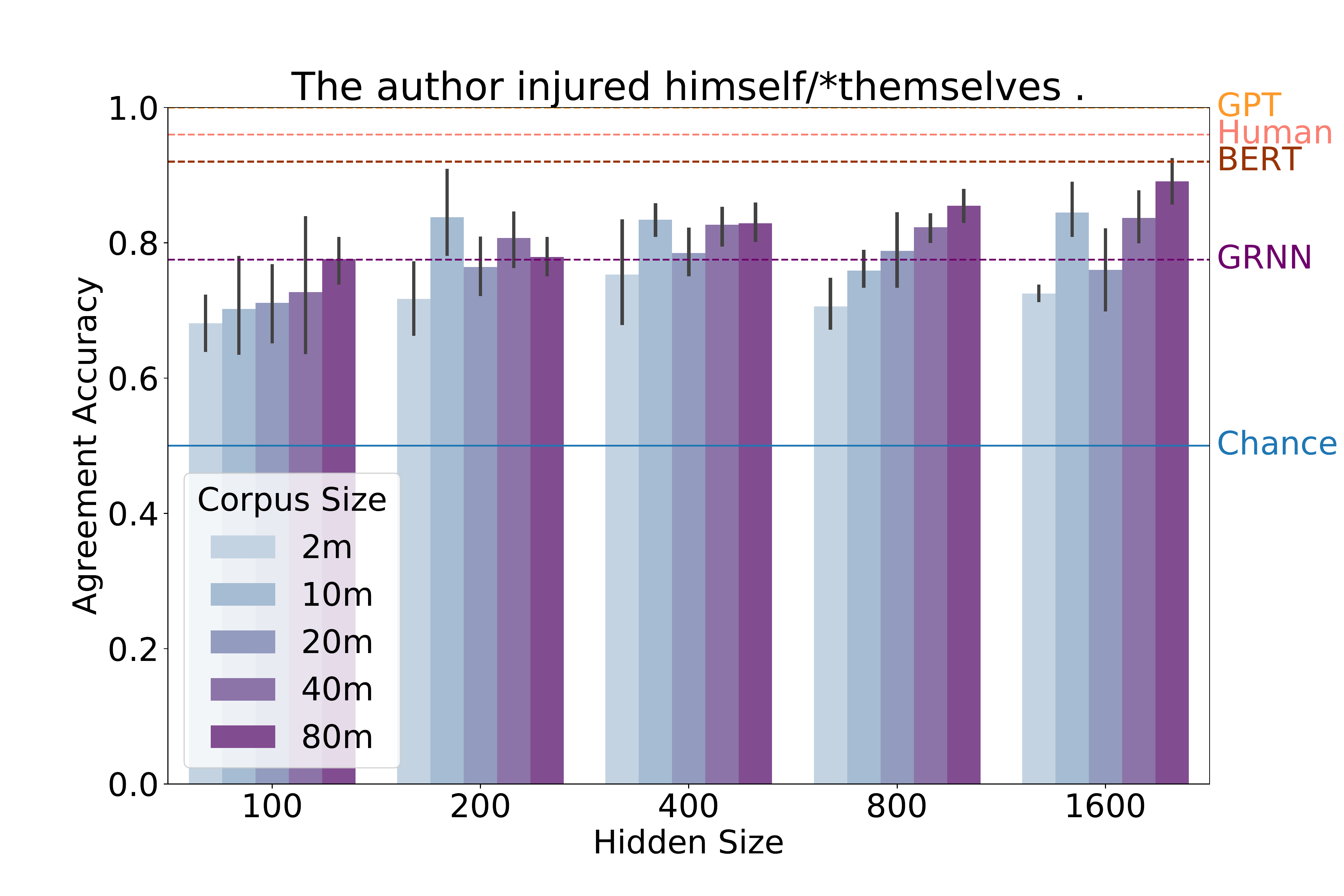}
  \caption{\label{fig:11} Reflexives: Simple}
  \end{subfigure}\hfill
  \begin{subfigure}[b]{0.5\textwidth}
  \includegraphics[width=\textwidth]{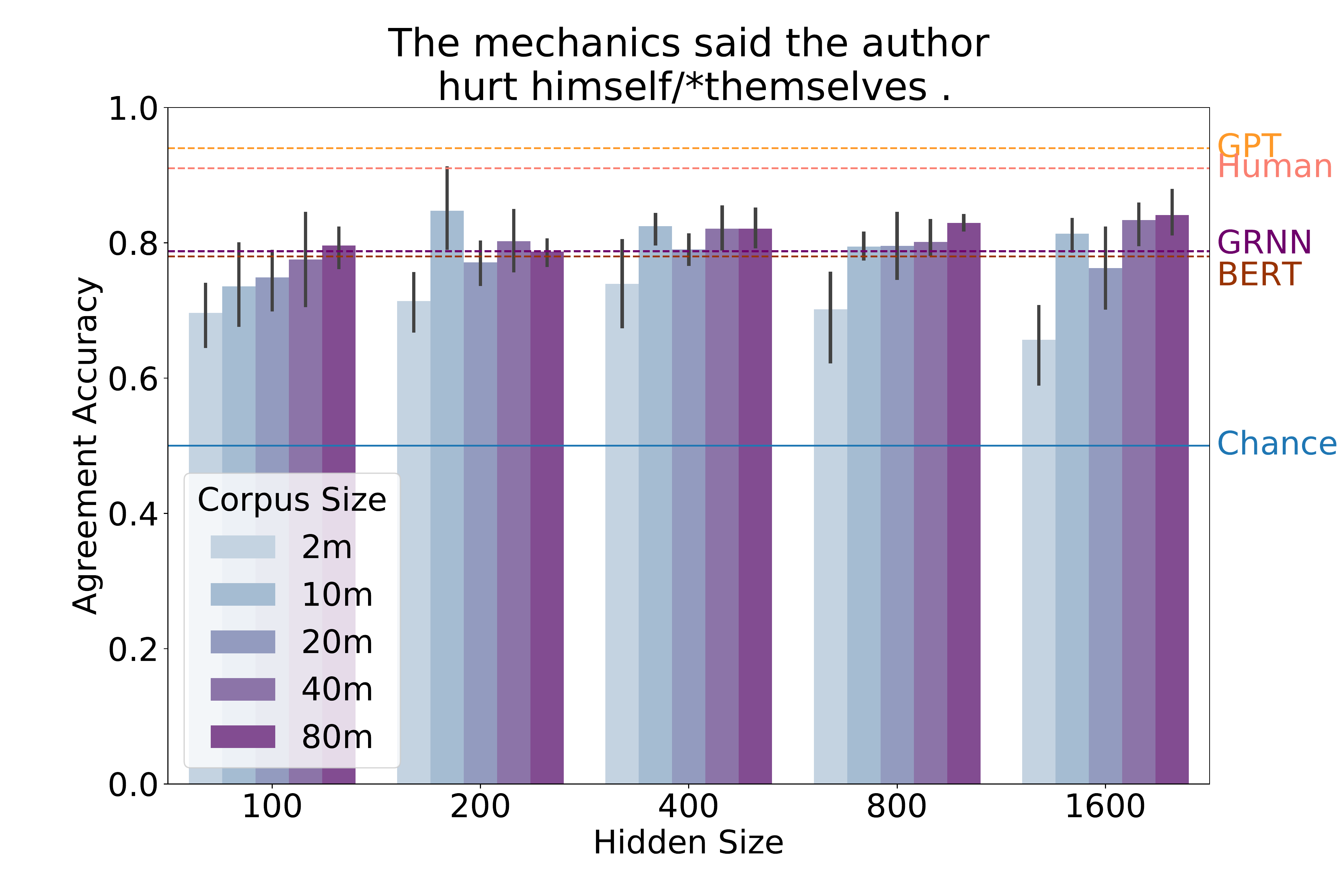}
  \caption{\label{fig:12} Reflexives: Sentential Complement}
  \end{subfigure}\\
  \begin{subfigure}[b]{0.5\textwidth}
  \includegraphics[width=\textwidth]{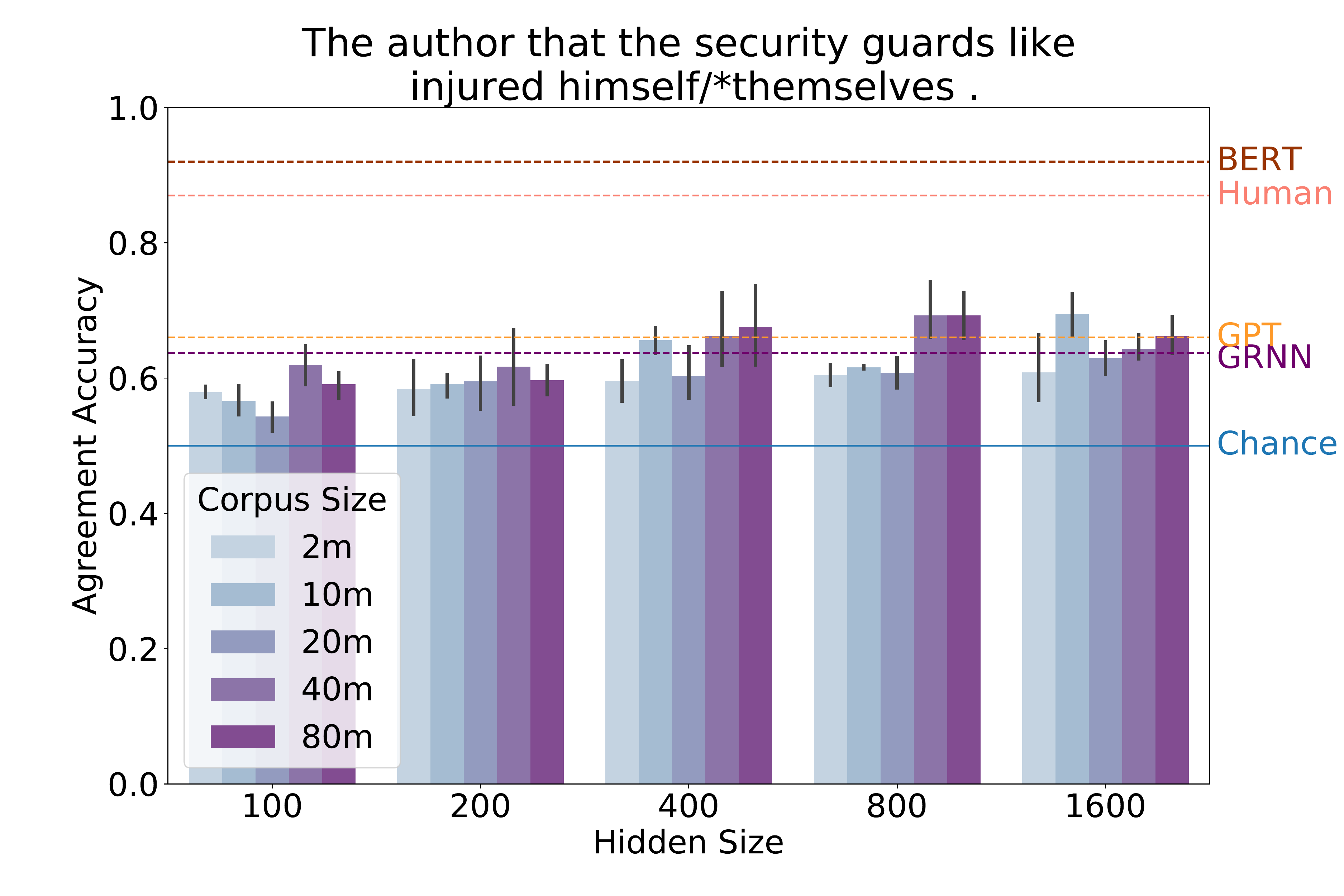}
  \caption{\label{fig:13} Reflexives: Across}
  \end{subfigure}
  \caption{Language model agreement performance between reflexive pronouns and their antecedent in simple transitive sentences (\ref{fig:11}), when agreement occurs within a sentential complement (\ref{fig:12}), and when there is an intervening subject relative clause (\ref{fig:13}). The dashed horizontal lines show agreement performance of commonly-used large-scale models. Error bars reflect standard deviation across the five models in each category. GPT and BERT results are from those reported by \protect\citet{wolfe19}. Human results are those reported by \protect\citet{marvinlinzen18}.}\label{fig:app3}
\end{figure*}

\clearpage

\begin{table*}
  \begin{tabular}{lrrrr}
      \toprule
    & 2M$\rightarrow$10M & 10M$\rightarrow$20M & 20M$\rightarrow$40M & 40M$\rightarrow$80M \\
      \midrule
    VP Coordination (short) & 4.0 & 0.4 & 0.5 & 0.6\\
    VP Coordination (long) & $1.8e^3$ & 0.3 & 18.8 & 0.4\\
    Subject Relative & 26.2 & 0.9 & 45.1 & 1.6\\
    Prepositional Phrase & $1.0e^5$ & 0.9 & 11.7 & 2.2 \\\hline
    Object Relative: Within & $8.7e^5$ & 0.4 & 1.7 & 1.4 \\
    Object Relative: Across & 0.4 & 0.4 & 1.5 & 1.1 \\
    Object Relative: Within (no that) & $1.4e^6$ & 0.8 & 0.5 & 1.4 \\
    Object Relative: Across (no that) & 0.9 & 0.4 & 28.8 & 0.4 \\\midrule
    Reflexives: Simple & 69.4 & 1.0 & 5.5 & 1.6\\
    Reflexives: Sentential Complement & 539 & 1.0 & 1.5 & 0.5\\
    Reflexives: Across & 18.3 & 6.4 & 9.7 & 0.4\\\bottomrule
  \end{tabular}
  \caption{Strength of evidence of improvement in each construction produced by increasing the training data (averaged across model sizes). For this analysis, we excluded models with fewer than 400 hidden units (i.e. those with 100 or 200 hidden units) and which therefore might not make effective use of additional training data. Strength of evidence is quantified by Bayes factors. A Bayes factor $K < 1$ indicates that there is no difference between the two model groups, and $K > 10$ provides strong evidence that the model groups obtain different accuracies.}\label{tab:incremental}
  \end{table*}

    \begin{table*}
    \centering
  \begin{tabular}{lrrr}
      \toprule
    & 10M$\rightarrow$20M & 20M$\rightarrow$40M & 40M$\rightarrow$80M\\
      \midrule
    VP Coordination (short) & - & $9e^{31}$ & $1e^{22}$ \\
    VP Coordination (long) & $4e^{78}$ & $1e^{10}$ & $2e^{16}$ \\
    Object Relative: Across & $7e^{20}$ & $2e^{10}$ & $2e^{10}$ \\
    Object Relative: Within (no that) & $5e^{12}$ & $8e^{13}$ & $1e^{11}$ \\
    Object Relative: Across (no that) & $1e^{38}$ & $4e^{10}$ & $2e^{16}$ \\\midrule
    Reflexives: Simple & - & $6e^{11}$ & $2e^{13}$ \\
    Reflexives: Sentential Complement & - & $1e^{12}$ & $1e^{18}$ \\ 
    Reflexives: Across & - & $6e^{11}$ & $3e^{23}$ \\\bottomrule
  \end{tabular}
  \caption{Training tokens needed for LSTMs to achieve human-like  performance in each condition that does not presently reach human-like performance. Projections were obtained by assuming that doubling the data produces a constant rate of error reduction. Each column of the table shows the results from assuming a different rate of error reduction, estimated from the error reductions we actually observed from each of our training data doublings. Increases in error (negative improvement) for a given doubling are denoted with `-' since those would never produce human-like performance. These results demonstrate that the human-like performance data requirements we report in our paper are actually fairly low compared to other improvement rates we observed.}\label{tab:appenough}
\end{table*}

  \begin{table*}
  \centering
  \begin{tabular}{lrrr}
      \toprule
    & 10M$\rightarrow$20M & 20M$\rightarrow$40M & 40M$\rightarrow$80M\\
      \midrule
    VP Coordination (short) & - & $3e^{96}$ & $1e^{60}$ \\
    VP Coordination (long) & $>1e^{100}$ & $2e^{19}$ & $1e^{58}$ \\
    Subject Relative  & - & $7e^{11}$ & $8e^{13}$ \\
    Prepositional Phrase & - & $7e^{11}$ & $4e^{13}$ \\\hline
    Object Relative: Within & $2e^{39}$ & $3e^{17}$ & $3^{17}$ \\
    Object Relative: Across & $7e^{73}$ & $1e^{19}$ & $1e^{18}$ \\
    Object Relative: Within (no that) & $2e^{38}$ & $1e^{46}$ & $9e^{28}$ \\
    Object Relative: Across (no that) & $>1e^{100}$ & $1e^{23}$ & $2e^{59}$ \\\midrule
    Reflexives: Simple & - & $5e^{18}$ & $9e^{22}$ \\
    Reflexives: Sentential Complement & - & $2e^{23}$ & $3e^{48}$ \\
    Reflexives: Across & - & $4e^{26}$ & $8e^{88}$ \\\bottomrule
  \end{tabular}
  \caption{Training tokens needed for LSTMs to achieve 99.99\% accuracy in each condition that does not presently reach 99\% accuracy. Projections were obtained by assuming that doubling the data produces a constant rate of error reduction. Each column of the table shows the results from assuming a different rate of error reduction, estimated from the error reductions we actually observed from each of our training data doublings. Increases in error (negative improvement) for a given doubling are denoted with `-' since those would never produce better performance.}\label{tab:appenoughperfect}
\end{table*}

\end{document}